\documentclass[conference]{IEEEtran}
\IEEEoverridecommandlockouts
\usepackage{cite}
\usepackage{mdframed}
\usepackage{comment}
\usepackage{colortbl, adjustbox, url}
\usepackage{threeparttable}
\usepackage{booktabs}

\usepackage[ruled,vlined,nosemicolon]{algorithm2e}
\usepackage{hyperref}
\usepackage{amsmath}
\usepackage{cleveref}
\usepackage{listings}
\usepackage{amsmath,amssymb,amsfonts}
\usepackage{bm}
\usepackage{graphicx}
\usepackage{textcomp}
\usepackage{tikz}
\usetikzlibrary{calc}
\usepackage{pgfplots}
\pgfplotsset{compat=1.18}
\usepackage{xcolor}

\pgfplotsset{
  colormap={jet}{
    rgb(0cm)  =(0.00, 0.00, 0.70);
    rgb(1cm)  =(0.00, 0.20, 1.00);
    rgb(2cm)  =(0.00, 0.75, 1.00);
    rgb(3cm)  =(0.00, 1.00, 0.50);
    rgb(4cm)  =(0.50, 1.00, 0.00);
    rgb(5cm)  =(1.00, 0.90, 0.00);
    rgb(6cm)  =(1.00, 0.40, 0.00);
    rgb(7cm)  =(0.75, 0.00, 0.00);
  }
}
\usetikzlibrary{positioning, backgrounds, shapes.geometric, arrows.meta, calc, shapes.symbols, fit}

\definecolor{pastelGreen}{RGB}{220, 240, 220}
\definecolor{pastelOrange}{RGB}{255, 235, 210}
\definecolor{pastelPurple}{RGB}{235, 220, 245}

\newcommand{\drawdag}[3]{
    \begin{scope}[shift={(#1)}, scale=#2]
        \node[circle, draw, inner sep=1.2pt, fill=white] (a) at (0,0.3) {};
        \node[circle, draw, inner sep=1.2pt, fill=white] (b) at (-0.3,0) {};
        \node[circle, draw, inner sep=1.2pt, fill=white] (c) at (0.3,0) {};
        \node[circle, draw, inner sep=1.2pt, fill=white] (d) at (0,-0.3) {};
        \begin{scope}[>={Stealth[scale=0.4]}] #3 \end{scope}
    \end{scope}
}

\tikzset{
  bn_node_a/.style={
    draw, circle,
    minimum size=1cm, inner sep=0,font=\sffamily\scriptsize, align=center
  }
}
\tikzset{
bn_node_e/.style={
draw, circle,
minimum size=1cm, inner sep=0,font=\sffamily\tiny, align=center}}
\usepackage{subcaption}

\def\BibTeX{{\rm B\kern-.05em{\sc i\kern-.025em b}\kern-.08em
    T\kern-.1667em\lower.7ex\hbox{E}\kern-.125emX}}

\begin{document}

\title{Causal Modeling of Adverse Pregnancy Outcomes via Adaptive LLM Proposals}

\author{
\IEEEauthorblockN{\bfseries
        Kavimayil P. Komarasamy\IEEEauthorrefmark{2}\IEEEauthorrefmark{1}\thanks{\IEEEauthorrefmark{1}Equal contribution.}, 
        Saurabh Mathur\IEEEauthorrefmark{3}\IEEEauthorrefmark{1},
        Ameet Soni\IEEEauthorrefmark{4}, 
        David M. Haas\IEEEauthorrefmark{5}, \\
        Kristian Kersting\IEEEauthorrefmark{3}\IEEEauthorrefmark{6}\IEEEauthorrefmark{7}, 
        Sriraam Natarajan\IEEEauthorrefmark{2}
    }
\IEEEauthorblockA{
\IEEEauthorrefmark{2}The University of Texas at Dallas}
\IEEEauthorblockA{
\IEEEauthorrefmark{3}Technical University of Darmstadt}
\IEEEauthorblockA{
\IEEEauthorrefmark{4}Swarthmore College}
\IEEEauthorblockA{
\IEEEauthorrefmark{5}Indiana University School of Medicine}
\IEEEauthorblockA{
\IEEEauthorrefmark{6}Hessian Center for Artificial Intelligence (hessian.ai)}
\IEEEauthorblockA{
\IEEEauthorrefmark{7}German Research Center for AI (DFKI)}
% \IEEEauthorblockA{
%         Emails: \IEEEauthorrefmark{2}\{kxp230053, Sriraam.Natarajan\}@utdallas.edu, 
%         \IEEEauthorrefmark{3}\{saurabh.mathur, kristian.kersting\}@tu-darmstadt.de, \\
%         \IEEEauthorrefmark{4}asoni1@swarthmore.edu, 
%         \IEEEauthorrefmark{5}dahaas@iu.edu
% }
}

\maketitle

\begin{abstract}
Adverse Pregnancy Outcomes (APOs) such as preterm birth and gestational diabetes can have long-term consequences for both the mother and child, yet an understanding of their causes remains elusive. Causal discovery in this domain is especially challenging due to a paucity of data and incomplete domain knowledge. As a result, pure data-driven methods fail, and Large Language Model (LLM) outputs remain inconsistent or contradictory. We introduce a neurosymbolic framework for generating plausible causal hypotheses that iteratively combines the broad prior knowledge of LLMs with empirical scoring on data. Our method treats the LLM as an adaptive proposal distribution, generating hypotheses that are scored against empirical data; the resulting high-scoring graphs are then used to update the LLM's context, steering subsequent generations toward more promising regions of the hypothesis space. We evaluate our approach on a real-world clinical dataset for modeling APOs and their risk factors, comparing our results against an expert-constructed causal graph. Our method recovers all expert-validated edges and identifies additional plausible causal relations not previously listed by experts, potentially providing new insights for targeted interventions.
\end{abstract}


\section{Introduction}
Reliable clinical reasoning requires robust causal models~\cite{kuipers1984causal,mathur2025teaching}, yet these models are rarely available in complex domains such as obstetrics. Modeling Adverse Pregnancy Outcomes (APOs) such as preeclampsia, gestational diabetes, and preterm birth is particularly challenging because they arise from an interplay of diverse risk factors, including maternal demographics, family history, pre-existing conditions, and lifestyle factors. While current medical literature provides high-precision knowledge of isolated links, it remains incomplete regarding the overall causal dynamics underlying APOs~\cite{numom2b}. Moreover, data-driven causal discovery is limited by the paucity of high-quality obstetric data and the extreme difficulty of performing interventional studies in pregnant populations.

On the other hand, while Large Language Models (LLMs) can suggest causal links based on their vast training corpora, they cannot reason causally; they fail to distinguish genuine causation from mere association~\cite{zevceviccausal,mathur2025llms}. Further, their outputs are highly sensitive to prompts and can result in contradictory graphs that lack grounding in empirical evidence. Existing hybrid approaches attempt to mitigate this through the theory refinement paradigm~\cite{buntine1991theory,mathur2025llm}, where an initial hypothesis is refined via local edits to improve its fit against the data. However, these methods are sensitive to their starting points, often becoming trapped in local optima that depend entirely on the quality of the initial hypothesis.

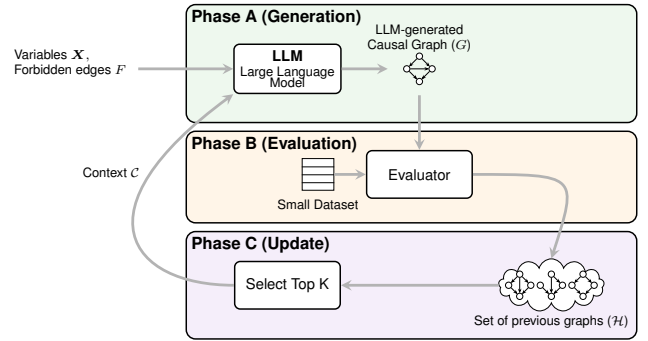
\begin{figure}[t!]
  
        \resizebox{.95\linewidth}{!}{\begin{tikzpicture}[
    font=\sffamily,
    % Reduced global node distance for compact layout
    node distance=0.8cm and 1cm,
    node_block/.style={draw, fill=white, rounded corners=3pt, minimum width=2cm, minimum height=0.9cm, align=center, thick, font=\footnotesize\sffamily},
    phase_box/.style={draw, fill=#1, rounded corners=5pt, thick}, % Parameterized color
    dataset/.style={draw, fill=white, minimum width=0.6cm, minimum height=0.6cm},
    main_arrow/.style={->, >=stealth, line width=1.5pt, black!30} % Neutral arrows for visibility
]

% 1. SYSTEMATIC GRID ANCHORS
\coordinate (LeftEdge)  at (-1.8, 0);
\coordinate (RightEdge) at (6.5, 0);
\coordinate (Col1)      at (0, 0);   % Processing column
\coordinate (Col2)      at (2.5, 0); % Distribution column
\coordinate (Col3)      at (5.0, 0); % Optimal result column

% 2. PHASE A (Generation) - Colored Pastel Green
\node[node_block] (LLM) at (Col1) {\textbf{LLM}\\[-1pt] \scriptsize Large Language\\[-3pt] \scriptsize Model};
\coordinate (G0_pos) at (Col2 |- LLM);
\node[above=0.2cm of G0_pos, font=\scriptsize\sffamily, align=center] (G0_text) {LLM-generated\\ Causal Graph ($G$)};

% Compacted DAGs
\drawdag{G0_pos}{0.8}{\draw[->](a)--(b); \draw[->](a)--(c); \draw[->](b)--(d); \draw[->](c)--(d); \draw[->](b)--(c);}

\begin{scope}[on background layer]
    \node[phase_box=pastelGreen!50, fit={(LeftEdge |- 0,1.1) (RightEdge |- 0,-0.9)}] (PhaseA) {};
\end{scope}
\node[anchor=north west, font=\small\bfseries\sffamily, inner sep=3pt] at (PhaseA.north west) {Phase A (Generation)};

% 3. PHASE B (Refinement) - Colored Pastel Orange
\node[node_block] (HC) at (Col2 |- 0,-2.0) {Evaluator};
\node[dataset, left=0.6cm of HC] (Data) {};
\foreach \i in {0.15,0.3,0.45} \draw ([yshift=\i cm]Data.south west) -- ([yshift=\i cm]Data.south east);
\node[below=1pt of Data, font=\scriptsize\sffamily] {Small Dataset};

\coordinate (Gstar_pos) at (Col3 |- HC);

\begin{scope}[on background layer]
    \node[phase_box=pastelOrange!50, fit={(LeftEdge |- 0,-1.3) (RightEdge |- 0,-2.8)}] (PhaseB) {};
\end{scope}
\node[anchor=north west, font=\small\bfseries\sffamily, inner sep=3pt] at (PhaseB.north west) {Phase B (Evaluation)};

% 4. PHASE C (Update) - Colored Pastel Purple
\node[node_block] (CF) at (Col1 |- 0,-4.1) {Select Top K};
\node[cloud, draw, fill=white, cloud puffs=13, minimum width=2cm, minimum height=1cm] (Cloud) at (Col3 |- CF) {};
\drawdag{Cloud.center}{0.7}{\draw[->](a)--(c); \draw[->](b)--(d); \draw[->](c)--(d); \draw[->](a)--(d);}
\drawdag{$(Cloud.center)+(0.6,0.0)$}{0.7}{\draw[->](a)--(b); \draw[->](a)--(c); \draw[->](b)--(d); \draw[->](c)--(d); }

\drawdag{$(Cloud.center)+(-0.6,0.0)$}{0.7}{\draw[->](a)--(b); \draw[->](a)--(c); \draw[->](b)--(d); \draw[->](c)--(d); \draw[->](a)--(d);}

\node[below=2.5pt of Cloud, font=\scriptsize\sffamily, align=center] {Set of previous graphs ($\mathcal{H}$)};

\begin{scope}[on background layer]
    \node[phase_box=pastelPurple!50, fit={(LeftEdge |- 0,-3.2) (RightEdge |- 0,-5.0)}] (PhaseC) {};
\end{scope}
\node[anchor=north west, font=\small\bfseries\sffamily, inner sep=3pt] at (PhaseC.north west) {Phase C (Update)};

\node[above left=-1.5cm and 1cm of PhaseA,align=left,font=\scriptsize\sffamily ] {Variables $\bm{X},$ \\Forbidden edges $F$};
\node[above left=-1cm and .75cm of PhaseB,align=left,font=\scriptsize\sffamily ] { Context $\mathcal{C}$};

% 5. CONNECTORS
\draw[main_arrow] ($(LLM.west)-(1.8,0)$) -- (LLM.west);
\draw[main_arrow] (LLM.east) -- ($(G0_pos)+(-0.6,0)$);
\draw[main_arrow] (Data.east) -- (HC.west);
\draw[main_arrow] (Cloud.west) -- (CF.east);
\draw[main_arrow] (G0_pos |- 0,-.5) -- (HC.north); 
\draw[main_arrow] (HC.east) .. controls +(2,0)  .. (Cloud.north);
\draw[main_arrow] (CF.west) .. controls +(-2.0,0) and +(-3,-2.2) .. (LLM.south west);

\end{tikzpicture}}
    
    \caption{\textbf{The \textsc{clara} Framework}. Our framework addresses causal discovery in data-sparse domains via a Generate-Evaluate-Update loop.
    It uses a pretrained LLM to {\em generate} diverse causal hypotheses, which are {\em evaluated} using a small clinical dataset. These scored graphs are used to identify patterns that consistently appear in top-scoring graphs and {\em update} the sampling distribution to shift mass towards more promising regions.}
    \label{fig:combined_methodology}
\end{figure}

In this work, we propose \textsc{clara} (Causal Learning via Adaptive Resampling and Aggregation), a neurosymbolic framework that treats the LLM as an adaptive proposal distribution, iteratively guided by data-driven evaluation. Inspired by the MIMIC stochastic optimization algorithm~\cite{bonet2021mimic}, \textsc{clara} iteratively samples causal structures from the LLM, scores them against empirical data, and updates the prompt for subsequent generations with the highest-scoring causal structures. This process effectively narrows the LLM's sampling mass toward high-confidence, data-validated structures.

Specifically, we make the following key contributions: (1) We introduce \textsc{clara}, a Neurosymbolic framework that combines approximate knowledge from an LLM with incomplete expert knowledge and a small empirical dataset to generate plausible causal hypotheses. (2) We demonstrate the efficacy of our method on a real-world clinical dataset for Adverse Pregnancy Outcomes. \textsc{clara} recovers all of the edges from an expert-constructed graph and uncovers additional, plausible causal relations beyond the expert graph. (3) We show how \textsc{clara} can be made more efficient without sacrificing performance.

The rest of the paper is organized as follows: after reviewing the necessary background and related work, we outline our \textsc{clara} algorithm and analyze it. We then present empirical evidence of this algorithm on the real problem of modeling adverse pregnancy outcomes before concluding the paper by outlining the areas of future research.

\section{Background}
\subsection{Causal Bayesian Networks} 
Causal Bayesian Networks (CBNs) are a class of causal models (CMs) and Probabilistic Graphical Models (PGMs)~\cite{koller2009probabilistic,pearl2009causality}. They represent causal relationships among a set of variables, $\bm{X}$ = $\{X_1,X_2, \dots, X_n\}$ using a causal graph $G = (\bm{X}, E),$ where each edge $X_i \rightarrow X_j \in E$ indicates that $X_i$ is a direct cause of $X_j.$ This differentiates a CBN from standard BNs that only encode probabilistic dependencies; CBNs encode causal relations and support interventional reasoning, which is critical for medicine. Formally, a CBN over $\bm{X}$ is defined as $\langle G, \bm{P} \rangle,$ where $G$ is the causal Directed Acyclic Graph (Causal DAG) and $\bm{P} = \{P_1,\dots,P_n\}$ is the set of local conditionals over each variable given its parents. The CBN factorizes the joint distribution over an assignment  $\bm{X} = \bm{x}$ as the product of local conditionals: $P(\bm{x}) = \prod_i P_i(X_i \mid \text{Pa}_i(\bm{x})).$

The focus of our work is generating plausible causal hypotheses: the task of identifying the graph ${G}$ that accurately represents the causal relations among variables $\bm{X}$. Data-driven methods such as Peter-Clark (PC~\cite{spirtes2001causation}), Fast Causal Inference (FCI~\cite{fci}), and Greedy Equivalence Search (GES~\cite{chickering2002ges}) learn the causal graph from observational data. They establish causation by combining patterns induced from large amounts of data with structural assumptions such as faithfulness, causal Markov condition, and causal sufficiency (see Section~\ref{subsec:dataset} for variable selection details). However, clinical datasets are typically small and noisy, with many relevant variables only weakly observed~\cite{ghassemi2020review}. Moreover, learning the optimal BN structure from data is difficult because the number of possible graph structures grows super-exponentially with the number of variables~\cite{chickering2004nphard}. As a result, algorithms that rely mainly on statistical signals such as conditional independence may fail to recover plausible causal structures in complex and data-scarce domains. 

\subsection{LLM-augmented Theory Refinement}
Theory refinement~\cite{mooney2021theoryrefinement} is a class of hybrid methods that address limitations of purely data-driven learning by exploiting domain knowledge. The use of domain knowledge allows theory refinement to reduce the search over a super-exponential number of possible graph structures to a local search in the neighborhood of an initial domain knowledge-based structure. This baseline BN structure is constructed using incomplete domain knowledge obtained from experts and is refined by local hill-climbing search; each step applies one of 3 types of atomic editing operations: adding, deleting, or reversing directed edges. Each operation is selected to maximize an empirical structure score, such as the Bayesian-Dirichlet (BD~\cite{bayesdirichlet}) score and the Bayesian Information Criterion (BIC~\cite{bic,mdl}). However, as the number of variables and potential relationships increases, specifying a good initial graph becomes challenging.

Recent work has addressed this issue using pretrained Large Language Models (LLMs~\cite{LLMsurvey}). These models have emerged as effective sources of {\it approximate domain knowledge} that can be used to create the initial hypothesis for theory refinement~\cite{petroni2019language,mathur2025llm}. These models capture causal domain knowledge from their vast training corpora, which include academic literature. However, while LLMs can generate fluent text, their reasoning capabilities remain limited, especially in causal inference~\cite{zevceviccausal}. Their outputs are stochastic and inconsistent across different runs. Since hill-climbing is a local search, relying on a single hypothesis from an LLM response may result in a structure that misses clinical nuances. 

\subsection{Estimation of Distribution Algorithms}
To overcome the sensitivity of theory refinement to poor initial hypotheses, we use Estimation of Distribution Algorithms (EDAs)~\cite{eda}. Unlike traditional optimization methods that track a single solution at a time, EDAs are a class of stochastic optimization methods that maintain and iteratively update a distribution over the entire search space. This population-based approach allows the algorithm to explore multiple neighborhoods of the search space simultaneously, reducing the risk of being trapped in local optima.

We specifically adopt the framework of Mutual-Information-Maximizing Input Clustering (MIMIC)~\cite{bonet2021mimic}, which provides a principled mechanism for optimization by transforming a simple initial distribution into one concentrated around optimal solutions through a Sample-Evaluate-Update loop. MIMIC draws samples from the distribution, evaluates them against the objective function, and updates the sampling distribution by learning the common structural features of the top-scoring samples. This process is repeated, refining the sampling distribution until it concentrates around a set of high-scoring candidate solutions.

\textsc{clara} instantiates this loop with two key substitutions. Rather than learning an explicit distribution from the top-scoring samples and drawing new candidates from it, \textsc{clara} uses an LLM as the sampling distribution, and updates subsequent generations by replacing its in-context examples with the top-scoring samples. This yields the Generate-Evaluate-Update loop described in Section~\ref{sec:method}.

\section{Causal Learning via Adaptive Resampling and Aggregation}\label{sec:method}
We aim to build a causal model by integrating approximate knowledge from an LLM, incomplete expert knowledge, and empirical evidence from data. We formalize this as the following problem:

\begin{mdframed}
    \textbf{Given:} A dataset $\mathcal{D}$ over discrete variables $\bm{X}=\{X_1, X_2, \ldots, X_n\}$,  incomplete expert knowledge in the form of a set of forbidden edges $\mathbf{F}$, and a pretrained Large Language Model $\mathcal{O}.$  \\\\
    \textbf{To Do:} Find a causal graph ${G}$ that accurately captures the causal relations over $\bm{X}$ by reconciling these knowledge sources.
\end{mdframed}
The primary challenge in this domain is the inherent incompleteness of information sources. While clinical experts provide high-precision knowledge, their expertise is often siloed within specific sub-specialties, leaving significant blind spots in the full causal structure. Further, the paucity of observational data prevents purely data-driven algorithms from fully capturing the causal structure, yielding overly sparse graphs. Finally, while Large Language Models (LLMs) act as broad knowledge sources, their outputs are fundamentally stochastic and unreliable; they typically fail to distinguish genuine causation from association and are highly sensitive to initial prompting, leading to unstable and inaccurate performance in one-shot causal graph generation.

To bridge this gap, we introduce Causal Learning via Adaptive Resampling and Aggregation (\textsc{clara}).
As outlined in Fig.~\ref{fig:combined_methodology}, \textsc{clara} instantiates a neurosymbolic system (Type 2: Symbolic[Neuro]~\cite{kautz2022third}) that finds causal graphs through a Sample-Evaluate-Update loop, gradually shaping the broad prior of a pretrained LLM to a distribution concentrated around data-validated plausible causal graphs.

\subsection{Sampling Causal Graphs}
\textsc{clara} starts by querying the LLM $\mathcal{O}$ for a batch of $K$ candidate causal hypotheses. This jumpstarts the search using the LLM's broad domain knowledge while mitigating the instability of committing to a single output. Each sample is obtained by prompting the LLM with the list of variables $\bm{X}$, along with brief descriptions, and expert knowledge in the form of forbidden edges $\bm{F}.$ Each candidate graph sampled from the LLM that violates forbidden edge and acyclicity constraints is discarded and resampled.

\subsection{Evaluating the Causal Graphs using Structure Score}
\textsc{clara} scores each LLM proposal $G'$ against the dataset $\mathcal{D}$ using the Bayesian Information Criterion (BIC)~\cite{bic} as a structure score. BIC balances the model's goodness of fit to the data against its complexity, and is defined as
\begin{equation}
    \text{BIC} (G') = \log P(\mathcal{D} \mid G') - \frac{\log N}{2} \sum_{i=1}^{n} q_i \left( |X_i| - 1 \right)
\end{equation}
\noindent where $\log P(\mathcal{D} \mid G')$ is the log-likelihood of the data $\mathcal{D}$ given graph $G'$, $N$ is the number of observations, $q_i=\prod_{X_j \in \text{Pa}(X_i)} |X_j|$ is the number of parent configurations of $X_i$, and $|X_i|$ is the number of states of $X_i$. In discrete Bayesian networks, each variable is parameterized by a conditional probability table (CPT) over all configurations of its parent variables, and the complexity penalty scales with the number of parameters. The complexity penalty term in BIC can massively penalize graph structures with a large number of parents and parameters, causing the score to underfit the data by favoring overly sparse graphs and pruning meaningful dependencies~\cite{Natarajan2006Structure}. To mitigate this, we account for local structure in the form of Context-Specific Independencies (CSI)~\cite{boutilier1996context} by representing each local conditional as a decision tree~\cite{friedman1996learning} instead of a full CPT. Each leaf stores a conditional distribution over the child variable conditioned on the parent assignments along the corresponding root-to-leaf path. This representation merges parent configurations when further splitting does not reduce the description length under the minimum description length criterion~\cite{mdl}. The resulting \textit{Tree BIC} score replaces the product of parent cardinalities in the Standard BIC penalty with the number of decision-tree leaves, 
\begin{equation}
\text{TreeBIC}(G') = \log P(\mathcal{D} \mid G') - \frac{\log N}{2} \sum_{i=1}^{n} l(T_i) \left( |X_i| - 1 \right)
\end{equation}
\noindent where $l(T_i)$ is the number of leaves in the decision tree, yielding a smaller penalty for context-specific structure. The scored graphs are then added to the history set $\mathcal{H}$ and ranked by their Tree BIC scores, ensuring that only the top-scoring graphs contribute to the context for the next generation.

\subsection{Updating the Sampling Distribution} Finally, to update the LLM sampling distribution for the next round of generation, \textsc{clara} constructs an in-context representation $\mathcal{C}$ from the best-scoring graphs in $\mathcal{H}$ to condition the next LLM generation, effectively narrowing the sampling distribution toward high-confidence structures. Since \textsc{clara} prompts the LLM to output graphs as lists of directed edges, we use a compatible representation (adjacency lists) in the in-context examples to ensure consistent semantics between prompt and output. As an alternative to constructing $\mathcal{C}$ from the full high-scoring graphs, we consider a compressed representation consisting of the edges common to all high-scoring graphs. We refer to this as the common edges representation and evaluate it as a more efficient alternative in Section~\ref{sec:eval}. This completes \textsc{clara}'s generate–evaluate–update loop, ensuring that each generation is conditioned on structures supported by the LLM's broad prior and empirical data.

\begin{algorithm}[t!]
    \caption{\small Causal Learning via Adaptive Resampling and Aggregation \textsc{(CLARA)}}
\label{alg:clara}
\KwIn{Variables $\bm{X}$, Observational Data $\mathcal{D}$, Forbidden Edges $\bm{F}$, Large Language Model $\mathcal{O}$, Batch Size $K$, Iterations $T$}
\KwOut{Final causal graph $G_{\text{Final}}$}
$\mathcal{H} \gets \emptyset$ \tcp*{Initialize history set}
$\mathcal{C}_0 \gets \emptyset$ \tcp*{Initialize representation set}
\While{$|\mathcal{H}| < K$\tcp*{Generate initial graphs}}{
    $G' \gets \textsc{SampleFromLLM}(\mathcal{O}, \bm{X}, \bm{F}, \mathcal{C}_0)$\;
    $G \gets \textsc{EvaluateWithData}(G, \mathcal{D}, \bm{F})$\;
    $\mathcal{H} \gets \mathcal{H} \cup \{G\}$\;
}
\For{$t = 1$ \KwTo $T$}{
    $\bm{G}_{\text{TopK}} \gets \textsc{SelectTopK}(\mathcal{H}, K)$ \tcp*{Top $K$ graphs ranked by Tree BIC}
    $\mathcal{C}_t \gets \textsc{InduceRepresentation}(\bm{G}_{\text{TopK}})$ \tcp*{Full Graphs or common edges}
    $G' \gets \textsc{SampleFromLLM}(\mathcal{O}, \bm{X}, \bm{F}, \mathcal{C}_t)$\;
    $G \gets \textsc{EvaluateWithData}(G, \mathcal{D}, \bm{F})$\;
    $\mathcal{H} \gets \mathcal{H} \cup \{G\}$ \tcp*{Update history}
}
$\bm{G}_{\text{TopK}} \gets \textsc{SelectTopK}(\mathcal{H}, K)$\; 
\Return $\textsc{AggregateCandidates}(\bm{G}_{\text{TopK}})$ \tcp*{Return union of best candidates}
\end{algorithm}

\subsection{The \textsc{clara} Algorithm}
Algorithm~\ref{alg:clara} presents the \textsc{clara} algorithm. It finds plausible causal structures by adapting the MIMIC algorithm's Generate-Evaluate-Update loop. Traditional MIMIC~\cite{bonet2021mimic} iteratively learns a simple density estimator such as a chain-structured Bayesian network over a bit-string encoding of the solution space and samples new candidates from the learned distribution. In contrast, \textsc{clara} replaces this learning-and-sampling step with a pretrained LLM, leveraging its encoded domain knowledge while conditioning generation on an in-context representation $\mathcal{C}$ constructed from the top-scoring graphs. The algorithm initializes the history set $\mathcal{H}$ by sampling an initial batch of $K$ causal graphs using \textsc{SampleFromLLM}, without any in-context examples ($\mathcal{C}_0=\emptyset$). This initial set of graphs is scored using \textsc{EvaluateWithData}, which computes the TreeBIC score. The evaluated graphs and their scores are then stored in $\mathcal{H}$. 

By treating the LLM as an adaptive proposal distribution, \textsc{clara} implements the MIMIC Generate-Evaluate-Update optimization loop. At each iteration, \textsc{SelectTopK} identifies the $K$ highest-scoring graphs stored in  $\mathcal{H}$, \textsc{InduceRepresentation} updates $\mathcal{C}$ based on the current set of top-$K$ causal graphs, and \textsc{SampleFromLLM} generates a new candidate causal graph conditioned on this context. \textsc{clara} evaluates each new LLM proposal using \textsc{EvaluateWithData}, which computes the structure score of the proposed causal graph based on $\mathcal{D}$. The evaluated graph and its score are then appended to the history set $\mathcal{H}$ before influencing subsequent generations, rather than treating the raw LLM output as a final hypothesis. While the construction of an in-context representation $\mathcal{C}$ serves as a distribution update, the scoring process ensures that only high-scoring candidates are used to construct the next generation context. Finally, once all $T$ iterations are complete, the top-$K$ high-scoring graphs in $\mathcal{H}$ are aggregated using a union-based strategy to construct the final causal graph. The details of the aggregation and cycle-breaking procedure are described in Section~\ref{ss:method_baselines}. This Generate-Evaluate-Update loop allows \textsc{clara} to effectively address the one-shot limitation of standard LLM-based causal graph generation with a global optimization strategy, which we evaluate empirically in Section~\ref{ss:eval_metrics}.

\section{Empirical Evaluation}\label{sec:eval}

We aim to answer the following questions:\\

\textbf{Q1:} Does \textsc{clara}'s combination of LLM priors and search yield better causal hypotheses than purely LLM-based generation, pure structure search, and LLM-initialized local search?\\
\textbf{Q2:} Does \textsc{clara} yield better causal hypotheses than data-driven causal discovery methods in data-scarce and noisy domains?\\
\textbf{Q3:} Does replacing DAGs with structural summaries reduce prompt size while maintaining performance?\\
\textbf{Q4:} Does \textsc{clara} propose plausible causal hypotheses in the real obstetrics domain, closely matching domain knowledge?\\

\subsection{Dataset Description}
\label{subsec:dataset}

To answer these questions, we employ two datasets. First, we generated a synthetic dataset based on the ALARM benchmark, a Bayesian network originally developed for patient monitoring and alarm systems~\cite{alarm}. The network contains 37 nodes and 46 edges, and both the ground-truth causal graph and conditional probability tables (CPTs) are obtained from the bnlearn repository\footnote{\url{https://www.bnlearn.com/bnrepository/}}. We sampled 3,000 observations from this network based on the provided probability tables and introduced 20\% entry-wise categorical noise, where each variable value is independently replaced with a randomly selected valid state with probability 0.2 to simulate imperfect observational conditions.

Second, we used real clinical data from the nuMoM2b study (Nulliparous Pregnancy Outcomes Study: Monitoring Mothers-to-be)~\cite{numom2b}, a large-scale longitudinal study of first-time mothers with singleton pregnancies. We focused on variables collected during the first prenatal visit to model the causal impact of early-pregnancy risk factors on APOs. The set of risk factors comprises 9 variables across four clinically salient domains: {\bf demographics, family history, pre-existing conditions, and lifestyle factors.} We consider 4 adverse pregnancy outcomes: {\bf preeclampsia, new hypertension, gestational diabetes, and preterm birth} (both spontaneous and medically indicated). These variables were selected by our clinical experts to ensure causal sufficiency to the best of current clinical knowledge. To ensure the integrity of the causal hypotheses generation, we performed targeted preprocessing. We excluded data from subjects with pre-existing diabetes to isolate the mechanisms associated with gestational diabetes. Further, we excluded all data points containing missing values to ensure the structural search relied strictly on observed clinical signals. Following these steps, we retained a final dataset of 3,856 examples. The full list of variables and their frequency distributions is presented in Table~\ref{tab:data_summary_stats}.

\begin{table}[t!]
\centering
\definecolor{demographic}{RGB}{232, 240, 255}   % pastel blue
\definecolor{familyhist}{RGB}{255, 236, 232}    % pastel red
\definecolor{preexisting}{RGB}{232, 250, 237}   % pastel green
\definecolor{lifestyle}{RGB}{245, 232, 255}     % pastel purple
\definecolor{outcomes}{RGB}{255, 248, 225}      % pastel yellow
\caption{\textbf{Baseline characteristics and pregnancy outcomes.} Variables and their frequencies in our dataset.
Row colors indicate variable type:
{\setlength{\fboxsep}{1.5pt}%
\colorbox{demographic}{demographic},
\colorbox{familyhist}{family history},
\colorbox{preexisting}{pre-existing conditions},
\colorbox{lifestyle}{lifestyle factors},
\colorbox{outcomes}{pregnancy outcomes}}.
All risk factors were measured at the start of pregnancy.}
\begin{tabular}{lr}
\toprule
\textbf{Variable} & \textbf{\%} \\
\midrule
\rowcolor{demographic} Race & \\
\rowcolor{demographic} \quad Non-Hispanic Asian & 4.2  \\
\rowcolor{demographic} \quad Non-Hispanic Black & 11.2 \\
\rowcolor{demographic} \quad Non-Hispanic White & 67.8 \\
\rowcolor{demographic} \quad Hispanic           & 12.3 \\
\rowcolor{demographic} \quad Other              & 4.4  \\
\rowcolor{demographic} Age & \\
\rowcolor{demographic} \quad $\leq$21           & 15.9 \\
\rowcolor{demographic} \quad 21--35             & 77.0 \\
\rowcolor{demographic} \quad $>$35              & 7.0  \\
\rowcolor{demographic} BMI & \\
\rowcolor{demographic} \quad $\leq$18           & 1.3  \\
\rowcolor{demographic} \quad 18--25             & 54.1 \\
\rowcolor{demographic} \quad $>$25              & 44.6 \\
\midrule
\rowcolor{familyhist} Family history of hypertension (HTNHist)    & 45.4 \\
\rowcolor{familyhist} Family history of type 2 diabetes (DiabHist) & 20.7 \\
\midrule
\rowcolor{preexisting} High blood pressure (HiBP)              & 2.7  \\
\rowcolor{preexisting} Polycystic ovary syndrome (PCOS)        & 4.8  \\
\midrule
\rowcolor{lifestyle} Smoked tobacco in 3 months before pregnancy (Smoked)    & 14.9 \\
\rowcolor{lifestyle} Adequate physical activity  (PhysAc) & 66.1 \\ % threshold ($>$450 METs)
\midrule
\rowcolor{outcomes} Preeclampsia  (PreEc)                      & 5.9  \\
\rowcolor{outcomes} New hypertension (NewHTN)                   & 17.7 \\
\rowcolor{outcomes} Gestational diabetes (GDM)               & 3.8  \\
\rowcolor{outcomes} Preterm birth (PTB)                   & 7.7  \\
\bottomrule
\end{tabular}
\label{tab:data_summary_stats}
\end{table}

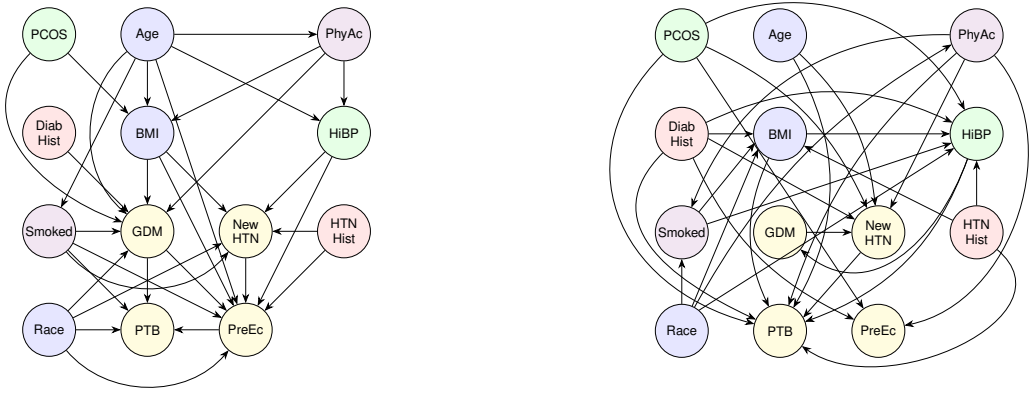
\begin{figure*}[t!]    
    \centering 
    \begin{subfigure}[t]{0.40\textwidth}
    \vspace{-2.5cm}
        \adjustbox{valign=c}{\scalebox{.7}{% In preamble, alongside your existing bn_node_a style:
\tikzset{
  bn_node_demo/.style={bn_node_a, fill=blue!10},
  bn_node_fam/.style={bn_node_a, fill=red!10},
  bn_node_pre/.style={bn_node_a, fill=green!10},
  bn_node_life/.style={bn_node_a, fill=violet!10},
  bn_node_out/.style={bn_node_a, fill=yellow!15},
}

\begin{tikzpicture}
      \draw
        (-0.929,  2.571) node[bn_node_demo] (Age){Age}
        (2.786,   0.714) node[bn_node_pre]  (HiBP){HiBP}
        (2.786,   2.571) node[bn_node_life] (METS){PhyAc}
        (-0.929,  0.714) node[bn_node_demo] (BMI){BMI}
        (-0.929, -1.143) node[bn_node_out]  (GDM){GDM}
        (0.929,  -3.0)   node[bn_node_out]  (PReEc){PreEc}
        (-2.786, -1.143) node[bn_node_life] (Smoking){Smoked}
        (0.929,  -1.143) node[bn_node_out]  (NewHTN){New\\HTN}
        (-2.786, -3)     node[bn_node_demo] (Race){Race}
        (-0.929, -3.0)   node[bn_node_out]  (PTB){PTB}
        (2.786,  -1.143) node[bn_node_fam]  (HTNHist){HTN\\Hist}
        (-2.786,  0.714) node[bn_node_fam]  (DiabHist){Diab\\Hist}
        (-2.786,  2.571) node[bn_node_pre]  (PCOS){PCOS};
      \begin{scope}[->, >=Stealth]
        \draw (Age) to (HiBP);
        \draw (Age) to (METS);
        \draw (Age) to (BMI);
        \draw (Age) to (PReEc);
        \draw (Age) to (Smoking);
        \draw (HiBP) to (PReEc);
        \draw (HiBP) to (NewHTN);
        \draw (METS) to (BMI);
        \draw (METS) to (GDM);
        \draw (METS) to (HiBP);
        \draw (BMI) to (GDM);
        \draw (BMI) to (PReEc);
        \draw (BMI) to (NewHTN);
        \draw (GDM) to (PReEc);
        \draw (GDM) to (PTB);
        \draw (PReEc) to (PTB);
        \draw (Smoking) to (GDM);
        \draw (Smoking) to (PTB);
        \draw (Smoking) to (PReEc);
        \draw (NewHTN) to (PReEc);
        \draw (Race) to (GDM);
        \draw (Race) to (PTB);
        \draw (Race) to (NewHTN);
        \draw (HTNHist) to (NewHTN);
        \draw (HTNHist) to (PReEc);
        \draw (DiabHist) to (GDM);
        \draw (PCOS) to (BMI);
        % \draw (PCOS) to (GDM);
      \end{scope}
      \path (Age) edge[->, >=Stealth, bend right=50] (GDM);
      \path (Smoking) edge[->, >=Stealth, bend right=50] (NewHTN);
      \path (Race) edge[->, >=Stealth, bend right=50] (PReEc);
      \path (PCOS) edge[->, >=Stealth, draw=black, out=225, in=160, looseness=1.2] (GDM);
    \end{tikzpicture}}}
    \end{subfigure} \hspace{0.02\textwidth} 
    \begin{subfigure}[t]{0.40\textwidth}
        \adjustbox{valign=c}{\scalebox{.7}{\tikzset{
  bn_node_demo/.style={bn_node_a, fill=blue!10},
  bn_node_fam/.style={bn_node_a, fill=red!10},
  bn_node_pre/.style={bn_node_a, fill=green!10},
  bn_node_life/.style={bn_node_a, fill=violet!10},
  bn_node_out/.style={bn_node_a, fill=yellow!15},
}

\begin{tikzpicture}
      \draw
        (-0.929,  2.571) node[bn_node_demo] (Age){Age}
        (2.786,   0.714) node[bn_node_pre]  (HiBP){HiBP}
        (2.786,   2.571) node[bn_node_life] (METS){PhyAc}
        (-0.929,  0.714) node[bn_node_demo] (BMI){BMI}
        (-0.929, -1.143) node[bn_node_out]  (GDM){GDM}
        (0.929,  -3.0)   node[bn_node_out]  (PReEc){PreEc}
        (-2.786, -1.143) node[bn_node_life] (Smoking){Smoked}
        (0.929,  -1.143) node[bn_node_out]  (NewHTN){New\\HTN}
        (-2.786, -3)     node[bn_node_demo] (Race){Race}
        (-0.929, -3.0)   node[bn_node_out]  (PTB){PTB}
        (2.786,  -1.143) node[bn_node_fam]  (HTNHist){HTN\\Hist}
        (-2.786,  0.714) node[bn_node_fam]  (DiabHist){Diab\\Hist}
        (-2.786,  2.571) node[bn_node_pre]  (PCOS){PCOS};

      \draw[->, >=Stealth]
        (Age)      edge [bend left=20] (NewHTN)
        (Age)      edge [bend left=30] (PTB)
        (BMI)      edge (HiBP)
        (BMI)      edge [bend right=25] (PTB)
        (DiabHist) edge (BMI)
        (DiabHist) edge [bend left=25] (HiBP)
        (DiabHist) edge [out=225, in=155, looseness=1.3] (PTB)
        
        (DiabHist.350) edge (NewHTN)
        (DiabHist) edge [bend right=20](PReEc)
        (GDM)      edge (NewHTN)
        (HTNHist)  edge (HiBP)
        % (HTNHist.225)  edge (PTB.-5)
        (HTNHist) edge [out=-40, in=325, looseness=1.2] (PTB)
        (HTNHist)  edge (BMI)
        (HiBP)     edge [bend left=25] (PTB)
        (HiBP)     edge [out=250, in=320, looseness=1.2] (GDM)
        %(HiBP.-130) -- (GDM)
        (METS)     edge (NewHTN)
        (METS)     edge [bend left=60](PReEc)
        (METS)     edge [bend left=-15]  (PTB)
        (METS)     edge [bend right=35](Smoking)
        (NewHTN)   edge (PTB.30)
        (PCOS)     edge [bend left=45] (HiBP)
        (PCOS)     edge [bend left=20] (NewHTN)
        (PCOS)     edge (PReEc)
        % (PCOS)     edge (PTB.135)
        (PCOS) edge [out=225, in=165, looseness=1.3] (PTB)
        (Race)     edge (BMI.215)
        (Race)     edge  (Smoking)
        (Race.50)   edge  (HiBP)
        (Race)     edge [bend left=15] (METS.200)
        (Smoking)  edge (HiBP.200)
        (Smoking.50) -- (BMI.200);

    \end{tikzpicture}}}
    \end{subfigure}
    \caption{\textbf{Expert-elicited causal graph} (left) and \textbf{additional edges discovered by \textsc{clara}} (right) for the obstetrics domain. Node color indicates variable type: {\setlength{\fboxsep}{1.5pt}\colorbox{blue!10}{demographic}}, {\setlength{\fboxsep}{1.5pt}\colorbox{red!10}{family history}}, {\setlength{\fboxsep}{1.5pt}\colorbox{green!10}{pre-existing conditions}}, {\setlength{\fboxsep}{1.5pt}\colorbox{violet!10}{lifestyle factors}}, {\setlength{\fboxsep}{1.5pt}\colorbox{yellow!15}{pregnancy outcomes}}. }
    \label{fig:apo_expert_graph}
\end{figure*}
\subsection{Method and Baselines}
\label{ss:method_baselines}
We compare \textsc{clara} with three types of baselines: (i) {\bf purely data-driven discovery}, (ii) {\bf one-shot LLM generation}, and (iii) {\bf LLM-augmented theory refinement}, as well as MIMIC~\cite{bonet2021mimic} as a non-LLM EDA ablation isolating \textsc{clara}'s iterative search loop from its LLM prior. The data-driven baselines are the PC~\cite{spirtes2001causation} and FCI~\cite{fci} algorithms to represent standard approaches for identifying causal structures from observational data. Both data-driven algorithms use the chi-squared conditional independence test with a significance level $\alpha=0.05$, consistent with the discrete-valued variables in both the ALARM and nuMoM2b datasets. LLM-dependent methods use GPT-5.2~\cite{chatgpt} and Llama-3.3-70B-Instruct(~\cite{meta2024llama33}) as the pretrained generators. Temperature is set to 0.7 across all LLM-based methods. The maximum token size is set to 8192 for Llama-3.3-70B-Instruct (other parameters default for both models). To reduce sampling variance, each candidate graph in \textsc{CLARA} is formed by the union of five independently sampled responses. Cycles introduced during aggregation are broken by removing the edge with the lowest frequency among five responses, with ties broken lexicographically (in ALARM, lexicographic resolution is used in 26\% of cycle-breaking events whereas no cycles occur in nuMoM2b). The forbidden edge set $\mathbf{F}$ encoding temporal order constraints as incomplete expert knowledge is available to all methods for nuMoM2b, whereas ALARM has no blacklist edges.

In addition, we evaluate two variants of \textsc{clara} to assess the impact of different components of the system: (1) \textsc{clara} (as fully implemented), using full graphs ranked by BIC score; (2) \textsc{clara} (Common edges), using representations enforcing direct adjacencies, included in-context in the prompt during the LLM-guided exploration. A common edge is included in the structural summary representation only if it appears in all top-K scoring graphs to reflect full agreement across the top-K set. For all experiments, we set the batch size K=7 and the number of iterations T=5, as this configuration achieves the lowest Structural Intervention Distance (SID)~\cite{peters2015sid} in the sensitivity analysis (Table~\ref{tab:k_t_analysis}), with no additional improvements observed from further iterations. Since LLM-based methods are inherently stochastic, all LLM-based methods were run five times independently, and results are reported as mean $\pm$ standard deviation across runs. Deterministic methods (PC, FCI) are reported as single values.

\subsection{Evaluation Metrics}
\label{ss:eval_metrics}
To evaluate structural correctness, we compare the learned causal graphs against a reference graph constructed by our obstetrics expert based on current medical consensus (Figure~\ref{fig:apo_expert_graph}). As obstetrics remains an active area of research, the expert-constructed causal graph in Figure~\ref{fig:apo_expert_graph} encodes all the causal relationships known to the best of our clinicians' knowledge. While we report {\bf Structural Hamming Distance (SHD)}~\cite{wahl2025shd}, {\bf Recall, and Precision} to measure topological alignment with this graph, we primarily focus on {\bf Structural Intervention Distance (SID)}~\cite{peters2015sid}. In high-stakes clinical domains like obstetrics, the functional reliability of a model in reasoning about the adjustment set for an intervention is more critical than exact adjacency recovery. We note that SHD treats every edge addition, deletion, and reversal as an equally severe error, regardless of whether that edge affects the adjustment set needed for a correct intervention. 

A graph can still have higher SHD while supporting correct interventional reasoning. \textit{For this reason, we treat SHD as a secondary metric and SID as our primary metric to support our comparative claims in discussion.} We further analyze the additional edges discovered beyond the expert reference based on whether they represent potential clinical discoveries.

\begin{table*}[t!]
\caption{\textbf{Comparative evaluation of \textsc{clara} and baselines on ALARM and nuMoM2b domains.}}
\label{tab:alarm_results}
\centering
\begin{tabular}{lcccc ccccc}
\toprule
& \multicolumn{4}{c}{\textbf{ALARM}} & \multicolumn{4}{c}{\textbf{nuMoM2b}} \\
\cmidrule(lr){2-5} \cmidrule(lr){6-9}
\textbf{Method} 
  & \textbf{SHD}~$\downarrow$ & \textbf{SID}~$\downarrow$ & \textbf{Recall}~$\uparrow$ & \textbf{Precision}~$\uparrow$
  & \textbf{SHD}~$\downarrow$ & \textbf{SID}~$\downarrow$ & \textbf{Recall}~$\uparrow$ & \textbf{Precision}~$\uparrow$ \\
\midrule
PC  & 35  & 314 & 0.5 & 0.6 & 34  & 83  & 0.1 & 0.3 \\
FCI & 46  & 368 & 0.1 & 0.5 & 32  & 79  & 0.0 & 0.0 \\
MIMIC & $66.2${\tiny$\pm 10.4$}  & $473.0${\tiny$\pm 15.4$}    & $0.0$ & $0.1$
& $37.6${\tiny$\pm 3.1$}  & $73.6${\tiny$\pm 2.5$}    & $0.1$ & $0.2$ \\
\midrule
\multicolumn{9}{l}{\textbf{LLM: GPT-5.2}} \\
\midrule
LLM (One-shot)
  & $48.2${\tiny$\pm 2.7$}  & $205.7${\tiny$\pm 18.0$} & $0.6$ & $0.4$
  & $29.2${\tiny$\pm 2.7$}  & $8.8${\tiny$\pm 4.4$}    & $1.0$ & $0.5$ \\
LLM + Theory Refinement
  & $41.6${\tiny$\pm 3.7$}  & $194.2${\tiny$\pm 12.6$} & $0.5$ & $0.5$
  & $27.4${\tiny$\pm 1.7$}  & $21.0${\tiny$\pm 0.0$}   & $0.8$ & $0.5$ \\
\textsc{clara}
  & $66.0${\tiny$\pm 4.1$}  & $139.8${\tiny$\pm 19.8$} & $0.7$ & $0.3$
  & $28.4${\tiny$\pm 1.2$}  & $0.0${\tiny$\pm 0.0$}    & $1.0$ & $0.5$ \\
\textsc{clara} (Common edges)
  & $67.8${\tiny$\pm 5.5$}  & $153.6${\tiny$\pm 24.0$} & $0.7$ & $0.3$
  & $29.0${\tiny$\pm 2.4$}  & $0.0${\tiny$\pm 0.0$}    & $1.0$ & $0.5$ \\
\midrule
\multicolumn{9}{l}{\textbf{LLM: Llama-3.3-70B-Instruct}} \\
\midrule
LLM (One-shot)
  & $74.4${\tiny$\pm 2.0$}  & $322.0${\tiny$\pm 33.0$} & $0.3$ & $0.2$
  & $26.8${\tiny$\pm 3.1$}  & $47.4${\tiny$\pm 3.7$}   & $0.4$ & $0.5$ \\
LLM + Theory Refinement
  & $66.6${\tiny$\pm 2.1$}  & $321.2${\tiny$\pm 27.6$} & $0.3$ & $0.2$
  & $26.4${\tiny$\pm 2.8$}  & $51.2${\tiny$\pm 4.1$}   & $0.4$ & $0.5$ \\
\textsc{clara}
  & $108.8${\tiny$\pm 4.0$} & $245.0${\tiny$\pm 30.3$} & $0.4$ & $0.1$
  & $26.4${\tiny$\pm 1.3$}  & $31.4${\tiny$\pm 3.3$}   & $0.6$ & $0.5$ \\
\textsc{clara} (Common edges)
  & $110.8${\tiny$\pm 5.9$} & $247.4${\tiny$\pm 23.8$} & $0.4$ & $0.1$
  & $25.8${\tiny$\pm 1.4$}  & $29.6${\tiny$\pm 4.8$}   & $0.6$ & $0.5$ \\
\bottomrule
\end{tabular}%}
\end{table*}

\subsection{Results and Discussion}
\label{subsec:results_discussion}
\textbf{(Answer, Q1:)} To answer Q1, we compared the causal networks generated by \textsc{clara} and the baselines to the ground-truth network in the synthetic ALARM dataset and an expert-constructed network in the real-world numom2b dataset. Table~\ref{tab:alarm_results} quantifies the divergence of the networks generated by each method with the reference network for each domain in terms of differences in structure (SHD, Precision, Recall) and differences in causal conclusions (SID). 

On the synthetic ALARM dataset, \textsc{clara} achieves the lowest SID across all methods, 32\% better than one-shot LLM generation, 55\% better than data-driven causal discovery, and 28\% better than the hybrid LLM-initialized theory refinement. This trend holds for both LLMs (GPT-5.2 and Llama-3.3-70B-Instruct), indicating that \textsc{clara}'s search procedure contributes to the improvement independent of the underlying LLM, though the magnitude of improvement varies by model. While LLM-generated networks captured more causal dependencies than purely data-driven causal discovery, they also captured more spurious relationships, increasing their recall at the cost of precision. Additionally, while LLM-initialized theory refinement eliminated some spurious edges from the LLM-generated network, its greedy search also eliminated weaker causal edges, increasing precision at the cost of recall. However, this tradeoff is unsuitable for high-stakes domains like medicine, where assuming causal independence is more costly than assuming causal dependence. \textsc{clara}  navigates this tradeoff more cautiously, capturing more causal dependencies by iteratively combining the LLM's broad prior with data-driven evaluation. Unlike theory refinement's greedy, one-edge-at-a-time pruning, CLARA's repeated resampling can retain edges that are individually weak but become informative jointly with other parents. 

The same pattern holds on the real-world nuMoM2b dataset (Table~\ref{tab:alarm_results}). \textsc{clara} achieves the lowest SID across all methods, outperforming one-shot LLM generation, data-driven causal discovery, and the hybrid LLM-initialized theory refinement across both LLMs. Moreover, in contrast to the synthetic ALARM domain, theory refinement eliminates a lot more causal edges, increasing the number of causal reasoning errors by 138\% relative to LLM One-shot. Providing the same forbidden edge set $\mathbf{F}$ to the data-driven baselines on nuMoM2b did not alter their scores, suggesting that \textsc{clara}'s improvements stem primarily from the LLM-derived priors. Consistent with this, MIMIC, which performs the same iterative resampling procedure without an LLM-informed prior, only marginally outperforms the purely data-driven baselines on nuMoM2b and yields the worst SID of any method on ALARM despite sharing \textsc{clara}'s architecture.

Overall, \textsc{clara} generates causal networks whose causal conclusions match the reference networks at a higher rate than purely LLM-based, data-driven causal discovery, and hybrid LLM-initialized theory refinement baselines. Therefore, we can answer {\bf Q1} affirmatively.

\begin{figure*}[t]
    \centering
    \includegraphics[width=.6\linewidth]{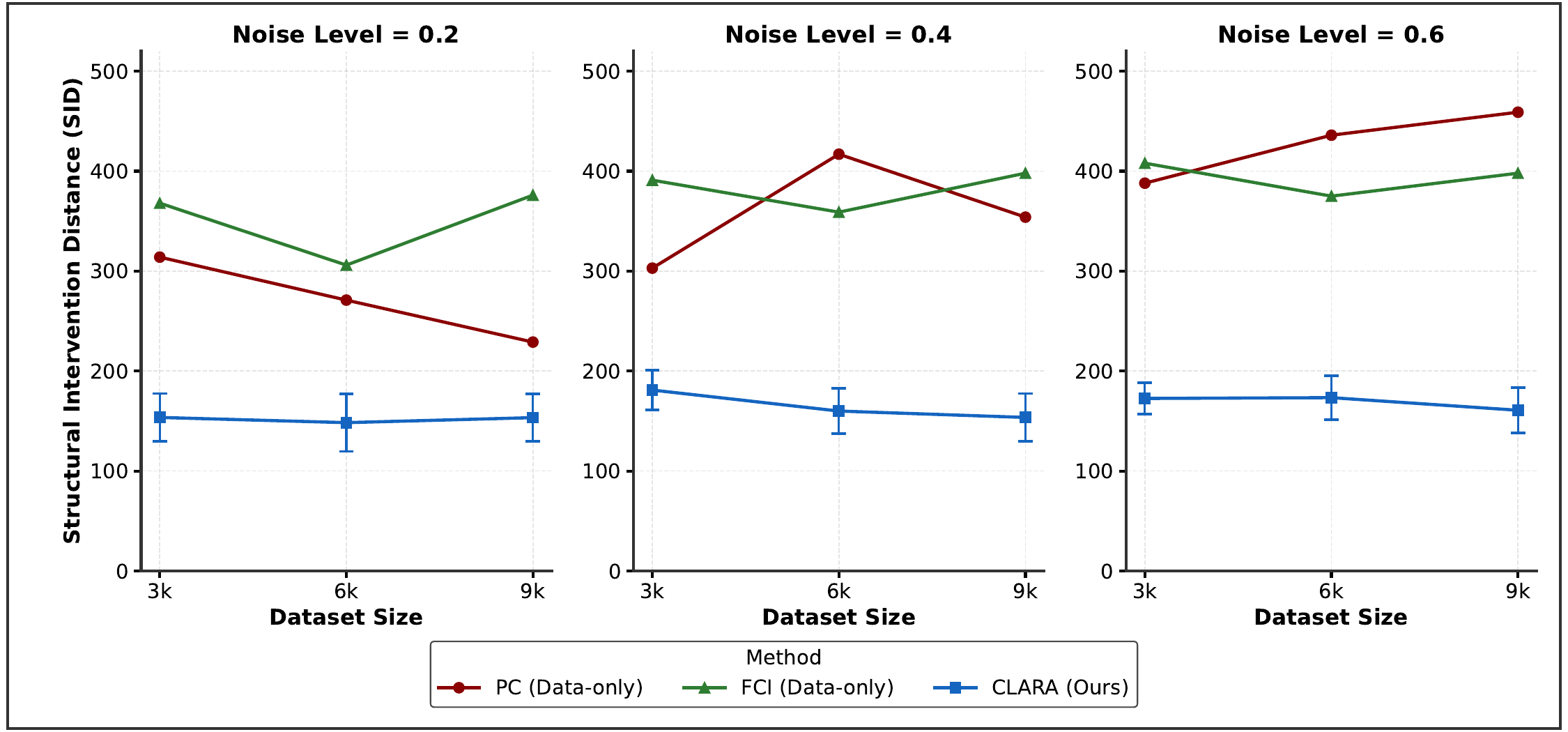}
    \caption{\textbf{Sensitivity of Structural Intervention Distance (SID) to noise and data size.} SID of PC, FCI, and CLARA across dataset sizes (3k,\,6k,\,9k) at noise levels 0.2,\,0.4,\,0.6. Each panel corresponds to a fixed noise level. PC and FCI are deterministic and reported as single-run values. \textsc{clara} SID values are the mean $\pm$ standard deviation across five independent runs (shown as error bars) using GPT-5.2 with K=7, T=5. Lower SID indicates fewer errors in predicted interventional distributions, i.e., better recovery of causal structure.}
    \label{fig:sid_dataset_noise_comparison}
\end{figure*}

\begin{table*}[h!]
\begin{minipage}{0.48\linewidth}
\caption{\textbf{Hyperparameter sensitivity analysis.} Mean and standard deviation of SID scores are reported for combinations of K (top-K best-scoring graphs) and T (number of iterations) on the ALARM dataset using GPT-5.2 as proposal LLM. Lower SID is better.}
\label{tab:k_t_analysis}
\setlength{\tabcolsep}{4pt}
\centering
\begin{tabular}{lcccc}
\toprule
\textbf{K/T} & 5                & 10                & 15                & 20                \\
\midrule
3 & $185.4${\tiny$\pm 25.9$} & $185.4${\tiny$\pm 25.9$} & $185.4${\tiny$\pm 25.9$} & $185.4${\tiny$\pm 25.9$} \\
5 & $176.0${\tiny$\pm 28.6$} & $176.0${\tiny$\pm 28.6$} & $173.2${\tiny$\pm 27.7$} & $173.2${\tiny$\pm 27.7$} \\
7 & $153.6${\tiny$\pm 24.0$} & $153.6${\tiny$\pm 24.0$} & $153.6${\tiny$\pm 24.0$} & $153.6${\tiny$\pm 24.0$} \\
\bottomrule
\end{tabular}
\end{minipage}
\hfill
\begin{minipage}{0.48\linewidth}
\caption{\textbf{Average prompt token counts by \textsc{clara} variant and dataset.} Average prompt size (tokens) per variant and dataset, measured across 5 independent runs for GPT-5.2 and Llama-3.3-70B.}
\label{tab:prompt_size}
\centering
\begin{tabular}{llcc}
\toprule
\textbf{Dataset} & \textbf{LLM}
& \textbf{Full}
& \textbf{Common edges} \\
\midrule
ALARM   & GPT-5.2                 & 3,825 & 1,377 \\
        & Llama-3.3-70B-Instruct  & 4,164 & 1,364 \\
\midrule
nuMoM2b & GPT-5.2                 & 2,535 & 1,195 \\
        & Llama-3.3-70B-Instruct  & 1,900 & 1,111 \\
\bottomrule
\end{tabular}
\end{minipage}
\end{table*}

\textbf{(Answer, Q2:)} To answer Q2, we analyze Figure~\ref{fig:sid_dataset_noise_comparison}, which plots SID across dataset sizes and noise levels for PC, FCI, and \textsc{clara} on the synthetic ALARM domain. \textsc{clara} maintains a low and stable SID across all combinations of dataset size and noise level. Across noise levels, \textsc{clara}'s SID remains largely unchanged as the proportion of randomly perturbed entries increases from 20\% to 60\%. Across dataset sizes, \textsc{clara}'s SID remains similarly low even as the number of samples is reduced, showing little dependence on the amount of available data. In contrast, while PC narrows the gap with \textsc{clara} as data increases at low noise (20\%), this improvement disappears at higher noise levels (40\% and 60\%), where its SID becomes non-monotonic and degrades with an increase in dataset size at the highest noise level. FCI remains elevated and non-monotonic across all combinations of dataset size and noise level, showing no clear benefit from additional data at any noise level. Notably, PC and FCI fail to match \textsc{clara}'s SID even when the dataset size is tripled. This stability across both axes indicates that \textsc{clara} generates better causal hypotheses than purely data-driven methods in both data-scarce and noisy clinical settings. We can therefore answer {\bf Q2} affirmatively.

\textbf{(Answer, Q3:)} To answer Q3, we examine the performance versus prompt size tradeoffs by comparing \textsc{clara} with a variant that replaces the full list of top-scoring graphs in the prompt with structural summaries in the form of edges common to top-scoring graphs. This yields up to three times fewer prompt tokens as shown in Table~\ref{tab:prompt_size}. This variant achieves an SID closely matching \textsc{clara} on three of the four dataset--LLM combinations. On ALARM with GPT-5.2, SID of the common edges variant is worse than \textsc{clara} by 10\%, but it is still 21\% better than the other baselines. Overall, these results indicate that replacing full DAGs with structural summaries reduces prompt size while maintaining performance close to that of fully implemented \textsc{clara}. We can therefore answer Q3 affirmatively.

\textbf{(Answer, Q4:)} To answer Q4, we compare the edges discovered by \textsc{clara} with the expert-constructed graph. \textsc{clara} recovers all 31 expert-validated causal edges in Figure~\ref{fig:apo_expert_graph} and proposes 30 additional edges. To assess their validity, we computed a positive likelihood ratio (LR+) for each using data and presented them to our obstetric clinical expert for rating against current medical consensus. The clinician rated 26 (87\%) as established, and the remaining 4 (13\%) as plausible: PCOS$\to$PTB, HiBP$\to$GDM, DiabHist$\to$PTB (likely mediated rather than direct), and Race$\to$PhyAc (varying in strength by Race). We identify three patterns in the additional edges.

First, \textsc{clara} identifies causal relationships among risk factors themselves. \textsc{clara} proposes family history of diabetes and hypertension as contributing causes of pre-pregnancy BMI and high blood pressure, consistent with evidence from a UK population-based cohort~\cite{diabhisttobmi} and a Japanese longitudinal study~\cite{histtobmihibp}. \textsc{clara} further proposes race as a contributing cause of BMI~\cite{racetobmi}, physical activity~\cite{racetomets}, and smoking~\cite{racetosmoking}, relationships supported by US-based evidence linking racial disparities to socioeconomic determinants of lifestyle. \textsc{clara} also proposes PCOS as a contributing cause of high blood pressure, confirmed by a meta-analysis of 30 studies in reproductive-age women~\cite{pcostohibp}.

Second, several additional edges are direct connections whose literature-established relationships are typically mediated by variables absent from the dataset. The direct edges from maternal age~\cite{agetoptbwithmediator} and pre-pregnancy BMI~\cite{bmitoptbgirsen} to preterm birth are well-supported but partly explained by intermediate factors such as gestational diabetes or preeclampsia. \textsc{clara}'s prior knowledge recovers these reduced-form direct effects in the absence of the mediating variables. Similarly, the direct edges from PCOS to preeclampsia and preterm birth are consistent with a meta-analysis showing a threefold increased preeclampsia risk in women with PCOS~\cite{pcostopreec}.

Third, \textsc{clara} identifies a sequential complication cascade among APOs: gestational diabetes as a contributing cause of new hypertension, supported by evidence that GDM substantially increases the risk of hypertensive disorders of pregnancy through shared metabolic mechanisms~\cite{gdmtonewthn}, and new hypertension as a contributing cause of preterm birth, consistent with evidence from a large Chinese prospective cohort~\cite{newhtntoptb}.

These findings, validated by our obstetric clinical expert, indicate that \textsc{clara} proposes plausible causal hypotheses closely matching domain knowledge in obstetrics. We can therefore answer Q4 affirmatively.

\section{Conclusion}
We addressed the task of generating plausible causal hypotheses for Adverse Pregnancy Outcomes (APOs). To do so, we introduced \textsc{clara}, a neurosymbolic framework that reconciles the broad priors encoded in pretrained Large Language Models with the empirical rigor of data-driven structure scoring. By framing the causal discovery process as a Sample-Evaluate-Update loop, \textsc{clara} effectively mitigates the stochasticity and unreliability inherent in one-shot neural proposals while overcoming the limitations of data sparsity in complex clinical domains. Moving forward, there are several promising directions to enhance this paradigm. First, the framework could be extended by incorporating more expressive constraint classes, such as qualitative influence statements indicating whether a cause suppresses or reinforces its effect, and constraints capturing time-varying and other context-specific clinical dependencies. Second, to address the computational challenges of high-dimensional domains, a hierarchical version of the estimation of distribution algorithm could be developed to decompose the causal search space into manageable sub-networks. Third, a systematic empirical comparison of the current union-based aggregation strategy with alternative approaches, such as higher-order DAG aggregation methods, would provide a better understanding of the trade-offs between different aggregation schemes. Finally, exploring methods for fine-tuning the underlying language models to make them more specialized and reliable generators of causal hypotheses remains important future work. Overall, together with these directions, \textsc{clara} presents an effective framework for causal discovery in complex, data-scarce medical domains.

\section*{Acknowledgment}
The authors gratefully acknowledge the support from NIH awards R01HD101246 and R01NS133142, and the Cluster of Excellence ``Reasonable AI" funded by the German Research Foundation (DFG) under Germany’s Excellence Strategy, EXC-3057.

\bibliographystyle{IEEEtran}
\bibliography{refs}

@book{pearl2009causality,
  title={Causality},
  author={Pearl, Judea},
  year={2009},
  publisher={Cambridge university press}
}

@book{koller2009probabilistic,
  title={Probabilistic {Graphical} {Models}},
  author={Koller, Daphne and Friedman, Nir},
  year={2009},
  publisher={MIT press}
}

@article{mathur2025teaching,
  title={Teaching Clinical Reasoning Through Cause-and-Effect Thinking: A Framework for Modern Medical Education},
  author={Mathur, Arvind},
  journal={The Clinical Teacher},
  year={2025},
  publisher={Wiley Online Library}
}

@inproceedings{Natarajan2006Structure,
  title={Structure Refinement in First Order Conditional Influence Language},
  author={Sriraam Natarajan and Weng-Keen Wong and Prasad Tadepalli},
  booktitle={ICML Workshop on Open Problems in Statistical Relational Learning},
  year={2006}
}

@inproceedings{bonet2021mimic,
 author = {De Bonet, Jeremy and Isbell, Charles and Viola, Paul},
 booktitle = {Advances in Neural Information Processing Systems},
 editor = {M.C. Mozer and M. Jordan and T. Petsche},
 pages = {},
 publisher = {MIT Press},
 title = {MIMIC: Finding Optima by Estimating Probability Densities},
 volume = {9},
 year = {1996}
}

@book{spirtes2001causation,
  title={Causation, prediction, and search},
  author={Spirtes, Peter and Glymour, Clark N and Scheines, Richard},
  year={2000},
  publisher={MIT press}
}

@inproceedings{mathur2025llm,
  title={LLM-Guided Causal Bayesian Network Construction for Pediatric Patients on ECMO},
  author={Mathur, Saurabh and Singh, Ranveer and others},
  booktitle={AIME},
  year={2025},
  organization={Springer}
}

@inproceedings{friedman1996learning,
  title={Learning Bayesian networks with local structure},
  author={Friedman, Nir and Goldszmidt, Moises},
  booktitle={UAI},
  year={1996}
}

@inproceedings{boutilier1996context,
  title={Context-specific independence in Bayesian networks},
  author={Boutilier, Craig and Friedman, Nir and Goldszmidt, Moises and Koller, Daphne},
  booktitle={UAI},
  year={1996}
}

@inproceedings{fci,
  title={An anytime algorithm for causal inference},
  author={Spirtes, Peter},
  booktitle={AISTATS},
  pages={278--285},
  year={2001},
  organization={PMLR}
}

@inproceedings{mooney2021theoryrefinement,
  author       = {Raymond J. Mooney and
                  Jude W. Shavlik},
  editor       = {Andreas Martin and
                  Knut Hinkelmann and
                  Hans{-}Georg Fill and
                  Aurona Gerber and
                  Doug Lenat and
                  Reinhard Stolle and
                  Frank van Harmelen},
  title        = {A Recap of Early Work on Theory and Knowledge Refinement},
  booktitle    = {Proceedings of the {AAAI} 2021 Spring Symposium on Combining Machine
                  Learning and Knowledge Engineering {(AAAI-MAKE} 2021), Stanford University,
                  Palo Alto, California, USA, March 22-24, 2021},
  series       = {{CEUR} Workshop Proceedings},
  volume       = {2846},
  year         = {2021}
}

@article{numom2b,
  title={A description of the methods of the Nulliparous Pregnancy Outcomes Study: monitoring mothers-to-be (nuMoM2b)},
  author={Haas, David M and Parker, Corette B and Wing, Deborah A and Parry, Samuel and Grobman, William A and Mercer, Brian M and Simhan, Hyagriv N and Hoffman, Matthew K and Silver, Robert M and Wadhwa, Pathik and others},
  journal={AJOG},
  year={2015},
  publisher={Elsevier}
}

@article{peters2015sid,
  title={{Structural Intervention Distance for Evaluating Causal Graphs}},
  author={Peters, Jonas and B{\"u}hlmann, Peter},
  journal={Neural Computation},
  volume={27},
  number={3},
  year={2015}
}

@inproceedings{wahl2025shd,
  title={Separation-Based Distance Measures for Causal Graphs},
  author={Wahl, Jonas and Runge, Jakob},
  booktitle={International Conference on Artificial Intelligence and Statistics},
  pages={3412--3420},
  year={2025},
  organization={PMLR}
}

@misc{chatgpt,
  author       = {OpenAI},
  title        = {ChatGPT: {GPT-5.2} Language Model},
  year         = {2025},
  howpublished = {\url{https://openai.com/chatgpt}},
}

@misc{meta2024llama33,
  author       = {{Meta}},
  title        = {Llama 3.3 70B Instruct},
  year         = {2024},
  howpublished = {\url{https://huggingface.co/meta-llama/Llama-3.3-70B-Instruct}}
}

@article{kuipers1984causal,
  title={Causal reasoning in medicine: analysis of a protocol},
  author={Kuipers, Benjamin and Kassirer, Jerome P},
  journal={Cognitive Science},
  year={1984},
  publisher={Elsevier}
}

@article{zevceviccausal,
    title={Causal Parrots: Large Language Models May Talk Causality But Are Not Causal},
    author={Matej Ze{\v{c}}evi{\'c} and Moritz Willig and Devendra Singh Dhami and Kristian Kersting},
    journal={TMLR},
    year={2023}
}

@inproceedings{buntine1991theory,
  title={Theory refinement on Bayesian networks},
  author={Buntine, Wray},
  booktitle={UAI},
  year={1991},
  publisher={Elsevier}
}

@inproceedings{mathur2025llms,
  title={LLMs for Causal Reasoning in Medicine? A Call for Caution},
  author={Mathur, Saurabh and Singh, Ranveer and Skinner, Michael and Radivojac, Predrag and Haas, David M and Raman, Lakshmi and Natarajan, Sriraam},
  booktitle={Proceedings of the 13th ACM IKDD International Conference on Data Science},
  pages={164--172},
  year={2025}
}

@article{LLMsurvey,
  title={Large language models: A survey},
  author={Minaee, Shervin and Mikolov, Tomas and Nikzad, Narjes and Chenaghlu, Meysam and Socher, Richard and Amatriain, Xavier and Gao, Jianfeng},
  journal={arXiv preprint arXiv:2402.06196},
  year={2024}
}

@inproceedings{petroni2019language,
  title={Language Models as Knowledge Bases?},
  author={Petroni, Fabio and Rockt{\"a}schel, Tim and Riedel, Sebastian and Lewis, Patrick and Bakhtin, Anton and Wu, Yuxiang and Miller, Alexander},
  booktitle={EMNLP-IJCNLP},
  year={2019}
}

@article{chickering2002ges,
  title={{Optimal} {Structure} {Identification} {With} {Greedy} {Search}},
  author={Chickering, David Maxwell},
  journal={JMLR},
  year={2002}
}

@article{mdl,
  title={Modeling by shortest data description},
  author={Rissanen, Jorma},
  journal={Automatica},
  year={1978},
  publisher={Elsevier}
}

@article{bayesdirichlet,
  title={Learning Bayesian networks: The combination of knowledge and statistical data},
  author={Heckerman, David and Geiger, Dan and Chickering, David M},
  journal={Machine Learning},
  year={1995},
  publisher={Springer}
}

@article{ghassemi2020review,
  title={A review of challenges and opportunities in machine learning for health},
  author={Ghassemi, Marzyeh and Naumann, Tristan and Schulam, Peter and Beam, Andrew L and Chen, Irene Y and Ranganath, Rajesh},
  journal={AMIA Summits on Translational Science Proceedings},
  year={2020}
}

@article{chickering2004nphard,
author = {Chickering, David Maxwell and Heckerman, David and Meek, Christopher},
title = {Large-Sample Learning of Bayesian Networks is NP-Hard},
year = {2004},
publisher = {JMLR.org},
journal = {J. Mach. Learn. Res.}
}

@article{bic,
  title={Estimating the dimension of a model},
  author={Schwarz, Gideon},
  journal={The Annals of Statistics},
  year={1978},
  publisher={JSTOR}
}

@article{kautz2022third,
  title={The third AI summer: AAAI Robert S. Engelmore memorial lecture},
  author={Kautz, Henry},
  journal={Ai magazine},
  year={2022}
}

@article{eda,
title = {An introduction and survey of estimation of distribution algorithms},
journal = {Swarm and Evolutionary Computation},
author = {Mark Hauschild and Martin Pelikan},
year = {2011}
}

@InProceedings{alarm,
author="Beinlich, Ingo A.
and Suermondt, H. J.
and Chavez, R. Martin
and Cooper, Gregory F.",
title="The ALARM Monitoring System: A Case Study with two Probabilistic Inference Techniques for Belief Networks",
booktitle="AIME 89",
year="1989",
publisher="Springer Berlin Heidelberg"
}

@article{pcostopreec,
    author = {Boomsma, CM and others},
    title = {A meta-analysis of pregnancy outcomes in women with polycystic ovary syndrome},
    journal = {Human Reproduction Update},
    year = {2006}
}

@article{bmitoptbgirsen,
  title={Women's prepregnancy underweight as a risk factor for preterm birth: a retrospective study},
  author={Girsen, Anna and others},
  journal={BJOG},
  year={2016},
}

@article{diabhisttobmi,
  title={Family history of diabetes identifies a group at increased risk for the metabolic consequences of obesity and physical inactivity in EPIC-Norfolk: a population-based study},
  author={Sargeant, LA and others},
  journal={Int. J. Obes.},
  year={2000},
  publisher={Nature Publishing Group}
}

@article{histtobmihibp,
  title={Cross-sectional and longitudinal associations between family history of type 2 diabetes mellitus, hypertension, and dyslipidemia and their prevalence and incidence: Toranomon Hospital Health Management Center Study (TOPICS24)},
  author={Ikeda, Izumi and others},
  journal={Mayo Clin. Proc.},
  year={2025},
  organization={Elsevier}
}

@article{racetobmi,
  title={Socioeconomic status, smoking, alcohol use, physical activity, and dietary behavior as determinants of obesity and body mass index in the United States: findings from the National Health Interview Survey},
  author={Shaikh, Raees A. and others},
  journal={Int. J. MCH AIDS},
  year={2015}
}

@article{racetosmoking,
  title={Age and racial/ethnic disparities in prepregnancy smoking among women who delivered live births},
  author={Tong, Van T and others},
  journal={Preventing Chronic Disease},
  year={2011}
}

@article{pcostohibp,
  title={Risk of hypertension in women with polycystic ovary syndrome: a systematic review, meta-analysis and meta-regression},
  author={Amiri, Mina and others},
  journal={Reprod. Biol. Endocrinol.},
  year={2020},
  publisher={Springer}
}

@article{newhtntoptb,
  title={Impact of gestational hypertension and pre-eclampsia on preterm birth in China: a large prospective cohort study},
  author={An, Hang and others},
  journal={BMJ Open},
  year={2022},
  publisher={British Medical Journal Publishing Group}
}

@article{gdmtonewthn,
author = {Sullivan, Shannon D and others},
title = {Hypertension Complicating Diabetic Pregnancies: Pathophysiology, Management, and Controversies},
journal = {J. Clin. Hypertens.},
year = {2011}
}

@article{racetomets,
  title={Neighborhood characteristics favorable to outdoor physical activity: disparities by socioeconomic and racial/ethnic composition},
  author={Franzini, Luisa and others},
  journal={Health \& Place},
  year={2010},
  publisher={Elsevier}
}

@article{agetoptbwithmediator,
  title={Effect of maternal age on the risk of preterm birth: A large cohort study},
  author={Fuchs, Florent and others},
  journal={PLOS ONE},
  year={2018},
  publisher={PLOS}
}

\end{document}